\documentclass[preprint,runningheads]{llncs}

\usepackage{booktabs}
\usepackage{multirow}
\usepackage{threeparttable}
\usepackage{siunitx}
\usepackage{rotating}
\usepackage{xcolor}

\usepackage{amssymb}
\usepackage{booktabs}
\usepackage{makecell}
\usepackage{amsmath}
\usepackage{url}
\usepackage{subfigure}
\usepackage{xcolor}
\usepackage{multirow}
\usepackage{threeparttable}
\usepackage[T1]{fontenc}
\usepackage{graphicx}
\usepackage{hyperref}

\begin{document}
%
\title{Structured Multi-Criteria Evaluation of Large Language Models with Fuzzy Analytic Hierarchy Process and DualJudge}
%
\titlerunning{Multi-Criteria LLM Evaluation with DualJudge and FAHP}
%


\author{Yulong~He\inst{1} \and
Ivan~Smirnov\inst{2} \and
Dmitry Fedrushkov\inst{2}\and
Sergey~Kovalchuk\inst{2}\and
Ilya~Revin\inst{2}
}
\authorrunning{He et al.}
%
\institute{St. Petersburg State University, University emb., 7/9, St. Petersburg, Russia \and
ITMO University, Kronverkskiy av., 49, St. Petersburg, Russia}
\maketitle              

\begin{abstract}
Effective evaluation of large language models (LLMs) remains a critical bottleneck, as conventional direct scoring often yields inconsistent and opaque judgments. In this work, we adapt the Analytic Hierarchy Process (AHP) to LLM-based evaluation and, more importantly, propose a confidence-aware Fuzzy AHP (FAHP) extension that models epistemic uncertainty via triangular fuzzy numbers modulated by LLM-generated confidence scores. Systematically validated on JudgeBench, our structured approach decomposes assessments into explicit criteria and incorporates uncertainty-aware aggregation, producing more calibrated judgments. Extensive experiments demonstrate that both crisp and fuzzy AHP consistently outperform direct scoring across model scales and dataset splits, with FAHP showing superior stability in uncertain comparison scenarios. Building on these insights, we propose \textbf{DualJudge}, a hybrid framework inspired by Dual-Process Theory that adaptively fuses holistic direct scores with structured AHP outputs via consistency-aware weighting. DualJudge achieves state-of-the-art performance, underscoring the complementary strengths of intuitive and deliberative evaluation paradigms. These results establish uncertainty-aware structured reasoning as a principled pathway toward more reliable LLM assessment. Code is available at \url{https://github.com/hreyulog/AHP_llm_judge}.

\end{abstract}
\section{Introduction}

The rapid progress of large language models (LLMs) has created an urgent demand for reliable and scalable evaluation methodologies.
Recent research increasingly adopts LLM-as-a-Judge, where language models automatically assess generated outputs as a substitute for costly human evaluation.
Such approaches have demonstrated promising alignment with human judgments across diverse tasks \cite{chiang2023closer,emirtekin2025llmreview}.
However, despite their growing adoption, existing evaluation paradigms largely rely on a single judging strategy, which introduces systematic biases across domains. Also, this strategy is not robust for different the instructions and prompts. \cite{liu2023gevalnlgevaluationusing}.

The dominant paradigm employs holistic evaluation, where an LLM directly produces an overall score or preference through intuitive judgment.
Direct scoring and pairwise comparison are efficient but implicitly compress multiple evaluation dimensions into a single reasoning step.
Prior studies show that LLM evaluators are sensitive to prompting formats and reasoning styles, resulting in unstable performance across task types \cite{chiang2023closer}. Moreover, LLM-based evaluation approach is inconsistent for different scoring scale \cite{li2026gradingscaleimpactllmasajudge}.
In particular, holistic judgments may fail to capture structured reasoning signals required in logic-intensive domains such as mathematics, coding, and legal analysis.

An alternative paradigm introduces structured evaluation, which decomposes judgments into multiple criteria and aggregates them using formal multi-criteria decision-making (MCDM) frameworks such as the Analytic Hierarchy Process (AHP).
Recent work has explored integrating AHP with LLMs for specific evaluation and decision-support scenarios \cite{lu2024ahp,xie2024dsgram,wu2026doc2ahp,llmAHP2025}, suggesting that hierarchical reasoning can improve interpretability and alignment with human assessment.
Nevertheless, the effectiveness of structured hierarchical evaluation as a general paradigm for LLM-as-a-Judge remains insufficiently understood, and systematic comparisons with holistic judging across heterogeneous domains are largely missing.

In this work, we conducted a comprehensive empirical study on JudgeBench \cite{tan2025judgebench}, covering 17 categories spanning mathematical reasoning, programming, scientific knowledge, and soft-knowledge tasks.
Our results reveal a clear pattern: structured evaluation based on AHP consistently outperforms direct holistic judging in logic-intensive domains, demonstrating that explicit criterion decomposition improves evaluation reliability.
Moreover, we find that modeling judgment uncertainty further enhances performance.
By extending AHP with triangular fuzzy numbers to represent preference ambiguity, Fuzzy AHP (FAHP) achieves higher accuracy than traditional crisp AHP, indicating that incorporating uncertainty better reflects the inherently probabilistic nature of LLM-generated judgments. We also treated scoring scale design as a controllable component of the system and showed the impact of the method for different scales.

Motivated by the complementary strengths of holistic and structured evaluation, we further introduce DualJudge, a hybrid framework that combines two independent evaluation signals: a direct holistic score and a structured AHP-based score (crisp or fuzzy).
Rather than introducing complex interactions, DualJudge aggregates the two scores through simple averaging, yielding consistently improved robustness across domains.
Despite its simplicity, this hybrid strategy achieves the best overall performance on JudgeBench.

Our contributions are threefold:
\begin{itemize}
    \item We provide a systematic empirical comparison between holistic and structured evaluation paradigms within LLM-as-a-Judge across diverse domains.
    \item We demonstrate that AHP-based structured evaluation significantly outperforms direct judging baselines, and that FAHP further improves performance by modeling judgment uncertainty.
    \item We propose DualJudge, a simple hybrid averaging framework that combines complementary evaluation signals and achieves the strongest overall accuracy.
\end{itemize}

\section{Related Work}

\subsection{LLM-as-a-Judge}
The rapid proliferation of LLMs has necessitated scalable evaluation methodologies, leading to the widespread adoption of the \textit{LLM-as-a-Judge} paradigm \cite{emirtekin2025llmreview}. This approach substitutes costly human annotation with automated assessments, demonstrating promising alignment with human preferences across diverse tasks. The dominant strategy in this domain is \textit{holistic evaluation}, where an LLM generates a single overall score or preference through intuitive judgment. While computationally efficient, holistic methods implicitly compress multiple evaluation dimensions into a single reasoning step. Prior studies indicate that such evaluators are highly sensitive to prompting formats and reasoning styles, resulting in unstable performance and systematic biases, particularly in logic-intensive domains such as mathematics and coding.
To mitigate these limitations, recent research has explored \textit{structured evaluation} paradigms. Unlike holistic judging, structured approaches decompose complex judgments into multiple criteria and aggregate them using formal frameworks \cite{lu2024ahp,xie2024dsgram}. Thus, the widely used prompt-based evaluation solution G-Eval \cite{liu2023gevalnlgevaluationusing} interacts with predefined evaluation criteria and detailed evaluation steps to obtain the aggregate score. This decomposition forces the model to engage with specific quality dimensions systematically, enhancing interpretability and reducing the opacity inherent in black-box verdicts \cite{wu2026doc2ahp}. Although initial work suggests that hierarchical reasoning improves alignment with human assessment \cite{llmAHP2025}, systematic comparisons between holistic and structured paradigms across heterogeneous domains remain insufficiently understood. Our work addresses this gap by empirically validating the superiority of structured reasoning in logic-driven tasks.

\subsection{Analytic Hierarchy Process}

The AHP, originally proposed by Saaty~\cite{saaty1980analytic,SAATY1987161}, is a classical MCDM framework designed to decompose complex decision problems into hierarchical structures. AHP organizes decision elements into multiple levels, typically consisting of a goal layer, a criteria layer, and an alternative layer, enabling systematic reasoning through pairwise comparisons.

Given a set of criteria, decision makers construct a pairwise comparison matrix
$\mathbf{A} = (a_{ij}), \quad a_{ij} > 0$, where $a_{ij}$ denotes the relative importance of criterion $i$ over criterion $j$. The priority weight vector $\mathbf{w}$ is obtained from the principal eigenvector:
$$
\mathbf{A}\mathbf{w} = \lambda_{\max}\mathbf{w},
$$
where $\lambda_{\max}$ represents the maximum eigenvalue of $\mathbf{A}$. The consistency ratio (CR) is further computed to ensure logical coherence in human judgments.

Due to its interpretability and structured reasoning capability, AHP has been widely applied in evaluation, ranking, risk analysis, and resource allocation tasks. In recent years, structured evaluation paradigms inspired by AHP have also been adopted in automated assessment systems and LLM evaluation pipelines, where complex judgments are decomposed into fine-grained criteria \cite{lu2024ahp}.

However, classical AHP assumes precise numerical comparisons between criteria. In practice, human judgments---and particularly LLM-generated evaluations---often exhibit uncertainty, vagueness, and linguistic ambiguity. This limitation motivates extensions that incorporate uncertainty modeling.

\subsection{Fuzzy Analytic Hierarchy Process}

To address uncertainty in pairwise comparisons, researchers introduced fuzzy set theory~\cite{ZADEH1965338} into AHP, resulting in the FAHP~\cite{Buckley1985FAHP}. FAHP replaces crisp comparison values with fuzzy numbers, enabling decision makers to express preferences using approximate or linguistic judgments.

A common formulation represents comparisons using triangular fuzzy numbers (TFN)~\cite{dubois1980fuzzy}:
$$
\tilde{a}_{ij} = (l_{ij}, m_{ij}, u_{ij}),
$$
where $l_{ij}$, $m_{ij}$, and $u_{ij}$ denote lower, modal, and upper bounds of preference intensity, respectively. Weight estimation is then performed through fuzzy aggregation and subsequent defuzzification procedures\cite{Buckley1985FAHP}.

Compared with standard AHP, FAHP provides several advantages:
\begin{itemize}
    \item improved robustness under uncertain or subjective judgments,
    \item better modeling of linguistic reasoning,
    \item reduced sensitivity to inconsistent comparisons.
\end{itemize}

These properties make FAHP particularly suitable for evaluation scenarios involving subjective assessment, incomplete information, or ambiguous criteria definitions. Consequently, FAHP has been increasingly applied in intelligent decision-support systems and complex evaluation environments.

\begin{figure}[t]
    \centering
    \includegraphics[width= \textwidth]{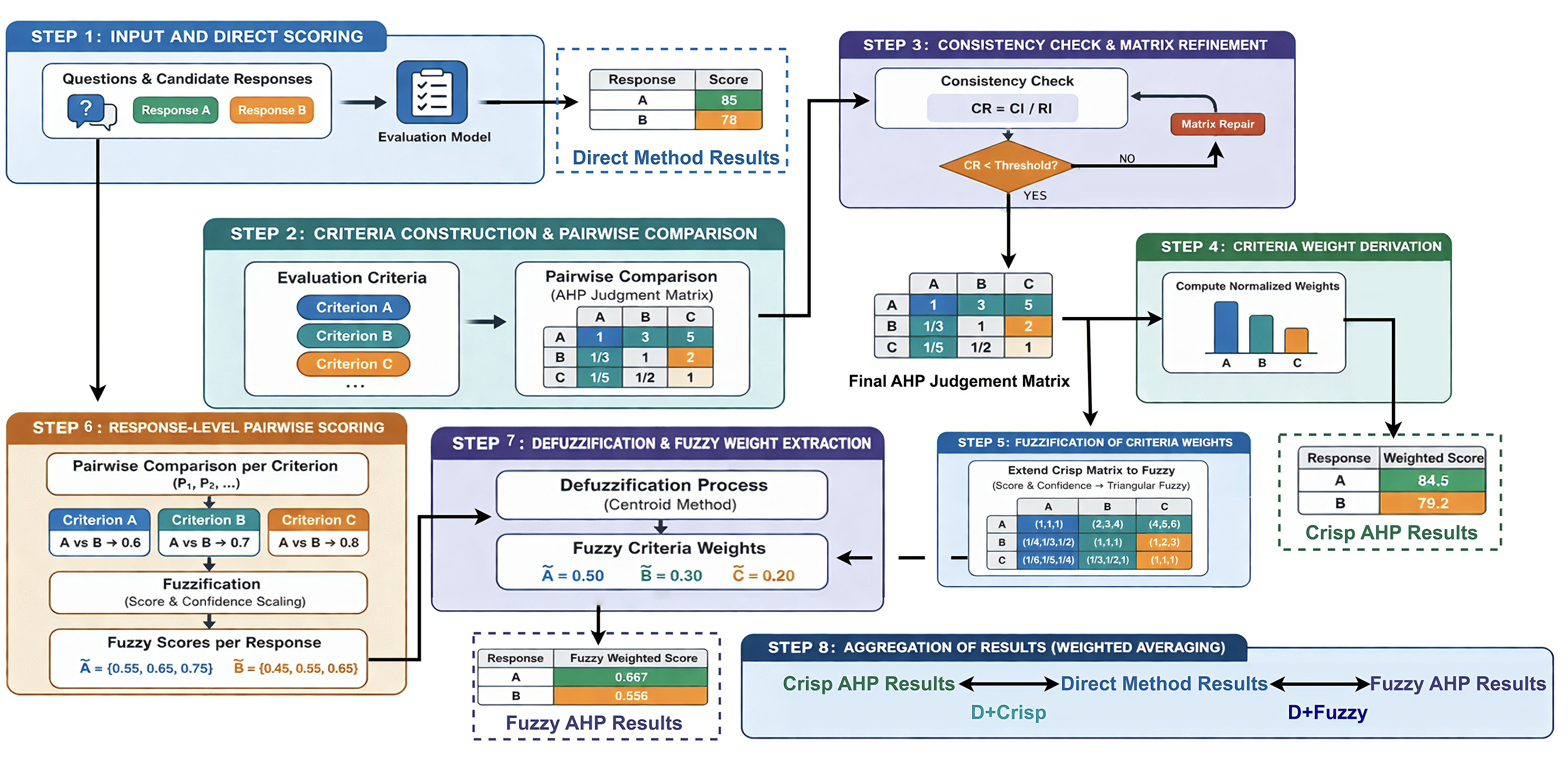}
    \caption{
    Overview of the proposed hybrid AHP-based evaluation framework.
    The pipeline integrates direct scoring, crisp AHP, and fuzzy AHP
    through consistency checking, weight derivation, and result aggregation.
    }
    \label{fig:framework}
\end{figure}

\section{Methodology}

This section presents the overall methodology of our study (see Fig ~\ref{fig:framework}), which aims to improve the reliability of LLM-based automatic evaluation through a hybrid judging framework. We first introduce the evaluation benchmark used in our experiments, followed by the structured decision-making mechanisms underlying our approach, and finally the proposed DualJudge framework. Our methodology is motivated by the observation that existing evaluation paradigms exhibit complementary strengths: intuitive scoring methods provide holistic judgments but lack interpretability, while structured approaches offer transparency and logical decomposition but may suffer from instability under uncertain reasoning. To address this limitation, we design a unified evaluation pipeline that integrates direct scoring with both crisp and fuzzy AHP formulations, and adaptively fuses their outputs based on consistency-aware reliability estimation. The resulting framework enables context-aware criterion construction, uncertainty modeling in pairwise comparisons, and adaptive aggregation of multiple evaluation signals, forming a principled hybrid judging process. The following subsections describe the dataset, the AHP-based evaluation mechanism, and the DualJudge architecture in detail.

\subsection{Dataset}
\label{subsec:dataset}

We evaluate our framework on \textbf{JudgeBench}~\cite{judgebench2024}, a benchmark specifically designed to assess the \textit{judgment capability} of LLMs. Unlike standard generation benchmarks that focus on output quality, JudgeBench constructs challenging pairwise comparison tasks: for each query, two model responses with minimal quality disparity are provided, along with a reliable ground-truth preference label. This design forces judge models to perform fine-grained, criteria-based analysis rather than relying on superficial heuristics (e.g., response length or formatting), making it an ideal testbed for investigating the limitations of single-paradigm evaluation methods.

\paragraph{Data Composition.}
JudgeBench aggregates questions from LiveBench~\cite{white2025livebenchchallengingcontaminationlimitedllm} and MMLU-Pro~\cite{wang2024mmluprorobustchallengingmultitask}, covering 17 fine-grained categories across STEM, humanities, and professional domains. To enable robust cross-generator validation, the benchmark provides two independent response sets:
\begin{itemize}
    \item \texttt{GPT Split}: Response pairs generated by GPT-4-class models (350 samples total).
    \item \texttt{Claude Split}: Response pairs generated by Claude-3-class models (270 samples total).
\end{itemize}
These splits ensure that observed improvements are not artifacts of a specific response style or generator bias. Table~\ref{tab:dataset_categories} details the category-wise distribution.

\begin{table}[t] 
\centering 
\small 
\caption{Category distribution of JudgeBench subsets by source and evaluation split. MMLU-Pro categories are aggregated for brevity.} 
\label{tab:dataset_categories} 
\begin{tabular}{l l c c} 
\toprule 
\textbf{Source} & \textbf{Categories} & \textbf{GPT} & \textbf{Claude} \\ 
\midrule 
\multirow{3}{*}{LiveBench} 
& livebench-math & 56 & 34 \\ 
& livebench-reasoning & 98 & 51 \\ 
& livecodebench & 42 & 31 \\ 
\midrule 
\multirow{1}{*}{MMLU-Pro} 
& 14 disciplines (math, physics, CS, law, etc.) & 154 & 154 \\ 
\midrule
\textbf{Total} & \textbf{17 categories} & \textbf{350} & \textbf{270} \\ 
\bottomrule 
\end{tabular} 
\end{table}

\subsection{Crisp and Fuzzy AHP Evaluation}
\label{sec:ahp_method}
\begin{figure}[t]
    \centering
    \includegraphics[width= 0.8\textwidth]{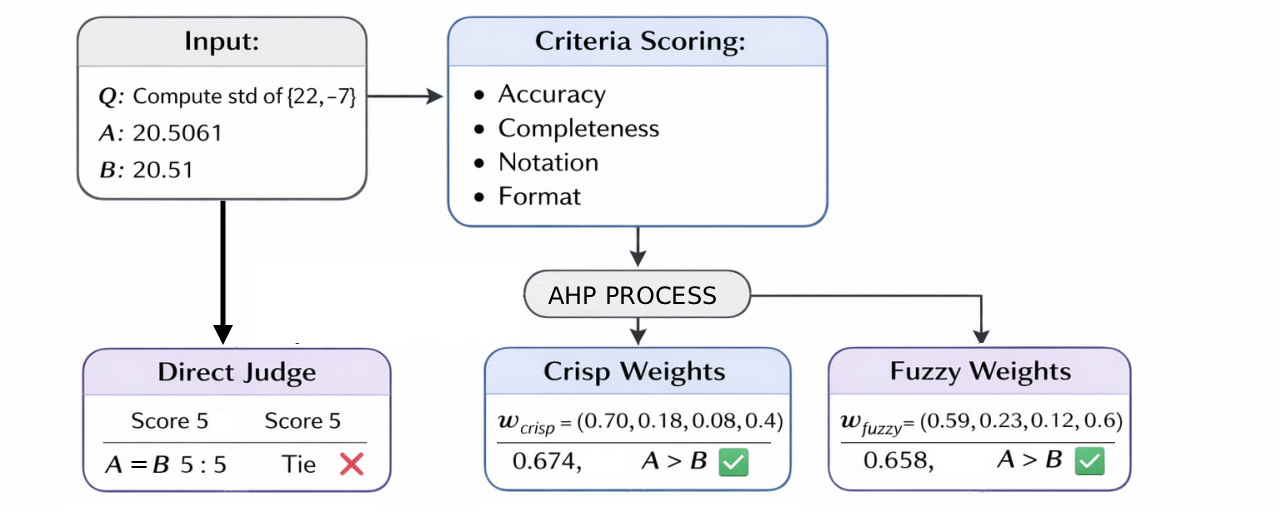}
    \caption{Despite a tie in direct scoring, the structured AHP-based evaluations
(crisp and fuzzy) consistently favor Response A.}
    \label{fig:case}
\end{figure}

Criterion importance is estimated using both the classical AHP and its uncertainty-aware extension, the FAHP. 
Both formulations operate on a shared pairwise comparison structure, differing only in how uncertainty in LLM judgments is modeled and propagated during aggregation. To provide further intuition into how the proposed framework operates in practice, we present a concrete example in Fig.~\ref{fig:case}.

\subsubsection{Pairwise Comparison Matrix Construction}

Let the evaluation criteria be
$\mathcal{C}=\{c_1,c_2,\dots,c_K\}.$
The LLM produces pairwise importance assessments represented as
$(c_i,c_j,s_{ij})$,
where $s_{ij}\in\{1,\dots,9\}$ follows Saaty's fundamental scale.

These comparisons define a reciprocal judgment matrix
\begin{equation}
\mathbf{A}=
\begin{bmatrix}
1 & a_{12} & \cdots & a_{1K}\\
1/a_{12} & 1 & \cdots & a_{2K}\\
\vdots & \vdots & \ddots & \vdots\\
1/a_{1K} & 1/a_{2K} & \cdots & 1
\end{bmatrix},
a_{ij} =
\begin{cases}
1, & i=j,\\
s_{ij}, & i<j,\\
1/s_{ji}, & i>j\\
\end{cases}
\end{equation}

\subsubsection{Crisp AHP Weight Estimation}

Criterion weights are computed using the principal eigenvector method.
Let $\lambda_{\max}$ denote the dominant eigenvalue of $\mathbf{A}$. 
The crisp weight vector $\mathbf{w}^c$ satisfies
$$
\mathbf{A}\mathbf{w}^c=\lambda_{\max}\mathbf{w}^c,
$$

followed by normalization
$$
w_i^c=\frac{v_i}{\sum_{k=1}^{K} v_k}.
$$

Logical consistency is quantified using the CR:
$$
CI=\frac{\lambda_{\max}-K}{K-1}, \qquad
CR=\frac{CI}{RI_K},
$$
where $RI_K$ denotes the random index for matrix size $K$.
The matrix is considered acceptable when $CR \le \tau$. where $\tau = 0.15$ ~\cite{Jos2006ConsistencyIT} is adopted as the consistency threshold.

\subsubsection{Automatic Consistency Repair}

When the consistency constraint is violated, a bounded automatic
repair procedure is applied to reduce local inconsistencies in the
judgment matrix. The repair enforces approximate multiplicative
transitivity,
$
a_{ik} \approx a_{ij}a_{jk},
$
which characterizes consistent pairwise comparisons in the analytic
hierarchy process \cite{saaty1980analytic}.

Entries exhibiting excessive logarithmic deviation from their
transitive estimates are locally adjusted while preserving the
reciprocity condition:$$
a_{ik}\leftarrow \hat{a}_{ik}, \qquad
a_{ki}=1/\hat{a}_{ik}.$$

The repair strategy is heuristic and performs a limited number of
local corrections within a single repair phase rather than iteratively
optimizing the entire matrix\cite{CRAWFORD1985387,ZESHUI1999443}. Updated entries are projected onto the
admissible Saaty scale range to preserve interpretability of comparison
intensities \cite{saaty1980analytic}.

After repair, the CR is recomputed. If the repaired
matrix satisfies the consistency requirement, it is accepted for weight
estimation. Otherwise, the judgment matrix is considered structurally
inconsistent and the pairwise comparisons are regenerated instead of
continuing further repair.

Matrices exhibiting unstable eigenvalue growth or extreme comparison
ratios are likewise treated as structurally inconsistent and trigger
regeneration of comparisons.

\subsubsection{Fuzzy AHP Representation}

To capture uncertainty inherent in LLM judgments, each discrete
comparison score is mapped to a TFN
$$
\tilde{a}_{ij}=(l_{ij},m_{ij},u_{ij}), \quad
l_{ij}\le m_{ij}\le u_{ij}.
$$

A predefined mapping function converts discrete scores into TFNs:
$$
\tilde{a}_{ij}=\mathrm{Scale}(s_{ij},\gamma_{ij}),
$$
where the confidence value $\gamma_{ij}\in[0,1]$ controls the
degree of uncertainty contraction.

Confidence directly shrinks the fuzzy interval toward the modal
value while keeping the modal judgment unchanged:
$$
l'_{ij}=l_{ij}+(m_{ij}-l_{ij})\gamma_{ij},\quad u'_{ij}=u_{ij}-(u_{ij}-m_{ij})\gamma_{ij}.
$$

The resulting confidence-adjusted TFN is $\tilde{a}'_{ij}=(l'_{ij},m_{ij},u'_{ij})$, such that higher confidence produces narrower fuzzy ranges, representing reduced epistemic uncertainty in LLM-generated
comparisons.

Reciprocity is preserved in the fuzzy domain:
$$
\tilde{a}'_{ji}=
\left(
\frac{1}{u'_{ij}},
\frac{1}{m_{ij}},
\frac{1}{l'_{ij}}
\right).
$$

\subsubsection{Fuzzy Weight Computation}

Criterion importance is estimated using the fuzzy geometric
mean method:
$\tilde{g}_i=
\left(\prod_{j=1}^{K}\tilde{a}_{ij}\right)^{1/K}.
$

Let $\tilde{g}_i=(l_i^{(g)},m_i^{(g)},u_i^{(g)})$ denote the
resulting TFN. Fuzzy weights are obtained via modal-based
scalar normalization:
$$
\tilde{w}_i=
\left(
\frac{l_i^{(g)}}{\sum_{k=1}^{K} m_k^{(g)}},
\frac{m_i^{(g)}}{\sum_{k=1}^{K} m_k^{(g)}},
\frac{u_i^{(g)}}{\sum_{k=1}^{K} m_k^{(g)}}
\right).
$$

Crisp priorities are obtained using centroid defuzzification:
$
w_i=\frac{l_i+m_i+u_i}{3},
$
followed by simplex normalization such that
$
\sum_{i=1}^{K} w_i=1.
$

\subsubsection{Shared Matrix Principle}

Both crisp and fuzzy weights are derived from the same repaired comparison matrix.
The fuzzy matrix is reconstructed from repaired scores while preserving confidence signals, ensuring that differences between AHP and FAHP arise solely from uncertainty modeling rather than inconsistent evidence.

\subsection{DualJudge}

We introduce \textbf{DualJudge}, an adaptive hybrid evaluation framework designed to address the limitations of single-paradigm LLM assessment. Existing approaches typically rely either on holistic, intuitive judgments (e.g., direct scoring) or structured, decompositional methods (e.g., AHP). Our empirical analysis suggests that these paradigms exhibit complementary error patterns across task domains.

DualJudge integrates both within a unified architecture conceptually inspired by Dual-Process Theory~\cite{kahneman2011thinking}, which distinguishes between fast, intuitive reasoning (System~1) and slow, deliberative reasoning (System~2). In our framework, the absolute scoring branch corresponds to intuitive evaluation, while the structured AHP branch reflects deliberative reasoning.

\paragraph{CR-Aware Adaptive Fusion.}
Let $S_{\text{ahp}} \in [0,1]$ and $S_{\text{abs}} \in [0,1]$ denote the evaluation scores produced by the structured and absolute branches, respectively. The final score is computed via an adaptive fusion mechanism.

We first incorporate a consistency screening step based on the AHP CR. Following standard AHP practice, structured evaluations with excessive inconsistency are considered unreliable. We therefore adopt the same consistency threshold $\tau$ 
defined in Section~3.2 and treat comparisons satisfying
$CR \le \tau$ as reliable.

Beyond binary filtering, we further interpret the CR as a continuous reliability signal~\cite{Saaty2012ModelsMC,ZESHUI1999443}. Lower CR values indicate stronger internal agreement among pairwise comparisons and thus higher confidence in the structured evaluation.

The fusion weight assigned to the structured branch is defined as
$$
\alpha(CR) = \exp(-\beta  CR),
$$
where $\beta$ controls the sensitivity to residual inconsistency. This formulation ensures that the structured branch maintains a non-negligible influence even under moderate inconsistency, reflecting the assumption that partially consistent structured reasoning still provides useful signals.

Since all observed CR values lie within the reliable regime, $\beta$ primarily modulates relative confidence rather than correcting severely inconsistent cases. We therefore fix $\beta = 7$ in all experiments.

The final score is computed as
$$
S_{\text{final}}=
\alpha(CR)S_{\text{ahp}}
+
\bigl(1 - \alpha(CR)\bigr)S_{\text{abs}}.
$$

\section{Experiments}

\subsection{Experimental Setup}

All experiments are conducted on JudgeBench using GPT-oss and Qwen3.5 \cite{qwen3.5} models as
the judge backbone. We evaluate two model scales for each model
(\texttt{gpt-oss-20b}, \texttt{gpt-oss-120b})~\cite{openai2025gptoss120bgptoss20bmodel} and \texttt{Qwen3.5-35B-A3B}, \texttt{Qwen3.5-9B} across both
\texttt{GPT} and \texttt{Claude} response splits to ensure robustness
to generator variation.

Two scoring granularities are considered (1--10 and 1--5 scales~\cite{li2026gradingscaleimpactllmasajudge}),
allowing us to analyze the effect of resolution on structured
evaluation stability.

Evaluation performance is measured as agreement accuracy with
ground-truth pairwise preferences. All LLM interactions are cached
to ensure reproducibility and eliminate sampling variance.

\subsection{Results}
\label{sec:experiments}
Table~\ref{tab:overall_summary} summarizes model performance across scales, scoring resolutions, while Tables~\ref{tab:gpt_combined_scales} (Appendix) provide detailed per-domain breakdowns.
\begin{table}[htbp]
\centering
\caption{Overall performance comparison across models and settings (Merged GPT/Claude splits)}
\label{tab:overall_summary}

\begin{tabular}{llcccccc}
\toprule
\multirow{2}{*}{\textbf{Model}} & \multirow{2}{*}{\textbf{Scale}} & \multirow{2}{*}{\textbf{Count}} & \textbf{Direct} & \multicolumn{2}{c}{\textbf{AHP}} & \multicolumn{2}{c}{\textbf{DualJudge}} \\
\cmidrule(lr){5-6} \cmidrule(lr){7-8}
& & & \textbf{(Baseline)} & \textbf{Crisp} & \textbf{Fuzzy} & \textbf{D+C} & \textbf{D+F} \\
\midrule
\multirow{2}{*}{gpt-oss-20b} & 1--10 & 620 & 69.83\% & 74.52\% & 75.35\% & 76.95\% & \textbf{77.60\%} \\
& 1--5  & 620 & 71.77\% & 75.48\% & 76.45\% & 78.06\% & \textbf{78.55\%} \\
\midrule
\multirow{2}{*}{gpt-oss-120b} & 1--10 & 620 & 75.81\% & 80.49\% & 80.97\% & \textbf{82.10\%} & \textbf{82.10\%} \\
& 1--5  & 620 & 75.00\% & 78.06\% & 78.23\% & \textbf{78.87\%} & 78.71\% \\
\midrule
\multirow{2}{*}{qwen3.5-9b} & 1--10 & 620 & 82.70\% & 81.44\% & 83.55\% & \textbf{84.07\%} & 84.03\% \\
& 1--5  & 620 & 81.91\% & 82.40\% & 83.87\% & \textbf{84.66\%} & 84.21\% \\
\midrule
\multirow{2}{*}{qwen3.5-35b} & 1--10 & 620 & \textbf{87.38\%} & 83.65\% & 85.47\% & 87.19\% & 87.19\% \\
& 1--5  & 620 & 85.73\% & 83.60\% & 86.24\% & \textbf{86.95\%} & 86.69\% \\
\bottomrule
\end{tabular}
\end{table}

\paragraph{Structured Reasoning vs. Direct Scoring.}
Across the aggregated results , structured evaluation methods (AHP and DualJudge) generally outperform direct scoring, with the magnitude of improvement inversely correlated with base model capability. For weaker evaluators like \texttt{gpt-oss-20b}, structured frameworks yield substantial gains: under the 1--10 scale, accuracy improves from 69.83\% (Direct) to 75.35\% (Fuzzy AHP) and peaks at 77.60\% (DualJudge D+F). The trend persists for \texttt{gpt-oss-120b} (+4.7--6.3 pp) and \texttt{qwen3.5-9b} (+1.4--2.0 pp). However, for the strongest evaluator \texttt{qwen3.5-35b} under the 1--10 scale, direct scoring (87.38\%) remains competitive, with structured variants achieving 85.47--87.19\%. This suggests that decomposed reasoning provides the most value when base judgments are noisy or poorly calibrated, while highly capable models may already internalize multi-criteria trade-offs.
\paragraph{Fuzzy vs. Crisp AHP.}
Fuzzy AHP consistently matches or exceeds Crisp AHP across model families and scales, supporting the hypothesis that explicit uncertainty modeling better accommodates the stochastic nature of LLM outputs. The advantage is most pronounced for mid-tier models: e.g., \texttt{gpt-oss-20b} (1--5 scale) improves from 75.48\% (Crisp) to 76.45\% (Fuzzy), and \texttt{qwen3.5-9b} (1--10) from 81.44\% to 83.55\%. For \texttt{qwen3.5-35b}, the gap narrows but Fuzzy still leads in 3 of 4 settings (e.g., 1--5 scale: 86.24\% vs. 83.60\%). The sole exception occurs for \texttt{qwen3.5-35b} (1--10), where both AHP variants underperform Direct scoring, indicating that fuzzy logic cannot fully compensate when the base evaluator's holistic judgments are already near-optimal.
\paragraph{DualJudge Performance.}
The hybrid DualJudge approaches (D+C and D+F) achieve peak accuracy in 7 of 8 configurations, demonstrating the benefit of fusing direct and structured signals. Notably, hybrid approaches show the most consistent gains on the \texttt{Claude} data split, which generally contains harder-to-distinguish response pairs \cite{judgebench2024}. Gains are largest for weaker models: \texttt{gpt-oss-20b} sees a +7.77 pp improvement (69.83\% → 77.60\%) under the 1--10 scale, while \texttt{gpt-oss-120b} gains +6.29 pp (75.81\% → 82.10\%). For stronger models, improvements diminish but remain positive: \texttt{qwen3.5-9b} gains +1.37--2.75 pp, and \texttt{qwen3.5-35b} shows marginal gains (+0.19 pp) or parity under 1--10. Notably, D+F (Fuzzy fusion) tends to lead for 20B-class models, while D+C (Crisp fusion) is competitive or superior for 9B/35B-class models, suggesting that the optimal fusion strategy may depend on the evaluator's inherent uncertainty calibration.
\paragraph{Scale Sensitivity and Task Generalization.}
The 1--5 scale generally yields slightly higher absolute accuracy than 1--10 across methods (e.g., \texttt{gpt-oss-20b} Direct: 71.77\% vs. 69.83\%), likely due to reduced granularity easing the judgment burden. However, the \textit{relative} gains from structured methods remain consistent across scales, indicating robustness to scoring protocol. Task-level analysis (see Appendix~\ref{sec:appendix_results}) reveals that gains concentrate on reasoning-intensive benchmarks: LiveCodeBench and MMLU-Pro mathematics/physics show 8--12 pp improvements for weaker models under DualJudge, while simpler factual categories exhibit smaller deltas. This pattern reinforces that structured decomposition is most valuable for cognitively complex, multi-attribute evaluation tasks.

Collectively, these aggregated results substantiate the central hypothesis: structured reasoning frameworks complement direct scoring, with adaptive fusion (DualJudge) yielding the most robust performance. The inverse relationship between base model capability and structured-method gains suggests a practical guideline: employ lightweight Direct scoring for state-of-the-art evaluators on well-defined tasks, but prioritize AHP/DualJudge pipelines for smaller models, ambiguous criteria, or high-stakes evaluation scenarios. The consistency across model families, scales, and evaluator backends (GPT vs. Claude splits in Appendix) underscores the generalizability of the proposed approach.

\section{Discussion}
\label{sec:discussion}
The empirical results presented in Section~\ref{sec:experiments} validate the core premise of our work: that hybrid evaluation frameworks can effectively mitigate the inherent instability of LLM-based judgment. The fact that DualJudge consistently outperforms other approaches across model scales and dataset splits suggests that combining fast, intuitive reasoning (System~1, based on the absolute scoring branch) with slower, deliberative reasoning (System~2, based on the structured AHP branch) creates something greater than the sum of its parts — the big-picture view of direct scoring covers for the blind spots of rigid step-by-step decomposition, and the other way around.
\paragraph{Structural Scaffolding for Weaker Evaluators.}
The magnitude of performance gains is inversely correlated with base model capability, indicating that structured reasoning acts as a cognitive scaffold. For \texttt{gpt-oss-20b}, DualJudge improves accuracy by +7.77 pp (1--10 scale), whereas \texttt{gpt-oss-120b} and \texttt{qwen3.5-9b} see gains of +6.29 pp and +1.37 pp, respectively. At the upper end, \texttt{qwen3.5-35b} already achieves 87.38\% with direct scoring, and structured variants yield parity or marginal changes. Weaker models, which often lack calibrated internal representations for complex evaluation tasks, benefit significantly from the explicit decomposition and consistency enforcement of AHP. Unlike standard chain-of-thought prompting, which relies on implicit step generation, our framework imposes mathematical constraints (e.g., consistency ratio checks) and explicit criterion weighting, providing a more rigorous scaffold for judgment calibration.

\paragraph{Modeling Epistemic Uncertainty.}
Fuzzy AHP consistently matches or exceeds Crisp AHP across model families and scales, underscoring the probabilistic nature of LLM outputs. LLMs do not produce deterministic truths; their token distributions reflect varying degrees of confidence. Crisp AHP forces these stochastic judgments into discrete Saaty scales, potentially discarding valuable nuance. By mapping pairwise comparisons to triangular fuzzy numbers (TFNs) modulated by confidence scores, Fuzzy AHP preserves uncertainty, allowing the aggregation mechanism to down-weight less reliable judgments. This advantage is most pronounced for mid-tier models and ambiguous tasks. However, when base evaluators are highly calibrated (e.g., \texttt{qwen3.5-35b}), the overhead of fuzzy decomposition yields diminishing returns, and direct scoring remains competitive. This suggests that future evaluation frameworks should adopt uncertainty-aware aggregation but may require adaptive routing to avoid unnecessary complexity for near-optimal baselines.

\paragraph{Domain Sensitivity and Adaptive Fusion.}
Performance gains are heterogeneous across domains, confirming that no single paradigm is universally optimal. On cognitively complex benchmarks like LiveCodeBench and MMLU-Pro mathematics/physics, DualJudge yields substantial improvements (e.g., +11.9 pp for \texttt{gpt-oss-20b} on LiveCodeBench), likely because multi-criteria decomposition aligns with the inherent structure of code correctness, efficiency, and style. In contrast, factual or recall-heavy domains (e.g., history, law) exhibit smaller deltas, as direct scoring already captures salient signals efficiently. Our adaptive fusion mechanism—which dynamically weights the structured branch based on consistency ratios and confidence—partially addresses this variability. Nevertheless, the results suggest that domain-aware routing or task-specific criterion weighting could further optimize the trade-off between evaluation fidelity and computational cost.

\paragraph{Practical Implications and Future Directions.}
Collectively, these findings offer actionable guidelines for LLM evaluation: (1) employ lightweight direct scoring for state-of-the-art models on well-defined tasks; (2) prioritize DualJudge pipelines for smaller models, high-stakes assessments, or ambiguous criteria; and (3) leverage fuzzy aggregation when evaluator calibration is uncertain or task complexity is high. Future work should explore dynamic thresholding for consistency checks, multimodal criterion decomposition, and real-time adaptive routing to minimize overhead while preserving robustness. As LLMs continue to scale, structured evaluation frameworks will remain essential for ensuring reliable, transparent, and reproducible judgment in automated assessment pipelines.

\section{Limitations and Future Work}
\label{sec:limitations}
While our results demonstrate the efficacy of the DualJudge framework, several limitations warrant acknowledgment and provide directions for future research.
\paragraph{Computational Overhead.}
The primary limitation of the proposed methodology is increased computational cost. The AHP component requires pairwise comparisons for criteria, and DualJudge necessitates running both direct and structured pipelines. For real-time evaluation or large-scale RLHF data processing, this latency may be prohibitive.
\textit{Future Work:} We plan to investigate efficiency optimizations, such as \textbf{sparse pairwise comparison} (evaluating only the most critical criterion pairs) or employing a \textbf{cascade strategy} where the lightweight Direct score is used first, and the expensive AHP pipeline is triggered only for low-confidence or high-stakes samples.
\paragraph{Dependency on Criteria Quality.}
Our experiments utilize a \textit{Category Mode} where shared criteria are derived for each dataset category. The performance of AHP is inherently bound to the relevance and completeness of these criteria. If the predefined criteria fail to capture key aspects of response quality, the structured evaluation may be systematically biased.
\textit{Future Work:} We aim to explore \textbf{dynamic criterion generation}, where the LLM generates task-specific evaluation criteria on a per-sample basis before constructing the pairwise matrix. This would enhance adaptability to novel or out-of-distribution tasks.
\paragraph{Model and Dataset Scope.}
Our study focuses on the \texttt{gpt-oss} and \texttt{qwen3.5} families and the JudgeBench benchmark. While JudgeBench is rigorous, it is limited to pairwise comparison tasks with ground truth. The generalizability to single-response scoring (e.g., reward modeling for RLHF) or open-weight models (e.g., Llama, Qwen) remains to be fully verified.
\textit{Future Work:} We intend to extend DualJudge to \textbf{reward modeling pipelines}, using the hybrid scores as fine-tuning signals for smaller reward models. Additionally, we will validate the framework across a broader spectrum of open-source architectures to ensure the findings are not model-specific.
\paragraph{Hyperparameter Sensitivity.}
The fusion mechanism relies on fixed hyperparameters. While these values performed robustly in our settings, optimal parameters may vary across different model families or task complexities.
\textit{Future Work:} Developing a \textbf{learnable fusion module} that adapts the weighting strategy based on task embeddings or model characteristics could further optimize performance without manual tuning.

\section{Conclusion}
\label{sec:conclusion}
In this work, we introduced DualJudge, a hybrid evaluation framework that unifies intuitive direct scoring with structured AHP reasoning. To validate the effectiveness of each component, we conducted a series of systematic comparative experiments across Direct Scoring, Crisp AHP, Fuzzy AHP, and the hybrid DualJudge framework. Our extensive experiments on JudgeBench demonstrate a clear performance hierarchy: structured methods (AHP/FAHP) consistently outperform direct scoring, with Fuzzy AHP further improving stability over Crisp AHP by explicitly modeling epistemic uncertainty. Ultimately, DualJudge achieves the highest accuracy by adaptively fusing these signals, yielding significant gains particularly in reasoning-intensive domains and for smaller model scales.
The findings underscore a critical insight for the field of automated evaluation: reliability is not solely a function of model scale, but also of reasoning architecture. By explicitly modeling the evaluation process through decomposed criteria and uncertainty-aware aggregation, we can extract more reliable signals from existing models. As LLMs are increasingly deployed in high-stakes decision-making scenarios, frameworks like DualJudge offer a principled path toward more transparent, robust, and trustworthy automated evaluation systems.

\bibliographystyle{splncs04}
\bibliography{custom1}

@misc{judgebench2024,
  title={JudgeBench: A Benchmark for Evaluating LLM-Based Judges},
  author={Sijun Tan and Siyuan Zhuang and Kyle Montgomery and Willian Y. Tang and Alejandro Cuadron and Chenguang Wang and Raluca Ada Popa and Ion Stoica},
  year={2024},
  archivePrefix={arXiv},
  url={https://arxiv.org/abs/2410.12784}
}

@book{kahneman2011thinking,
  title={Thinking, fast and slow},
  author={Kahneman, Daniel},
  year={2011},
  publisher={macmillan}
}

@article{SAATY1987161,
title = {The analytic hierarchy process—what it is and how it is used},
journal = {Mathematical Modelling},
volume = {9},
number = {3},
pages = {161-176},
year = {1987},
issn = {0270-0255},
doi = {https://doi.org/10.1016/0270-0255(87)90473-8},
url = {https://www.sciencedirect.com/science/article/pii/0270025587904738},
author = {R.W. Saaty}
}

@article{ZADEH1965338,
title = {Fuzzy sets},
journal = {Information and Control},
volume = {8},
number = {3},
pages = {338-353},
year = {1965},
issn = {0019-9958},
doi = {https://doi.org/10.1016/S0019-9958(65)90241-X},
url = {https://www.sciencedirect.com/science/article/pii/S001999586590241X},
author = {L.A. Zadeh}
}

@inproceedings{chiang2023closer,
    title = "A Closer Look into Using Large Language Models for Automatic Evaluation",
    author = "Chiang, Cheng-Han  and
      Lee, Hung-yi",
    editor = "Bouamor, Houda  and
      Pino, Juan  and
      Bali, Kalika",
    booktitle = "Findings of the Association for Computational Linguistics: EMNLP 2023",
    month = dec,
    year = "2023",
    address = "Singapore",
    publisher = "Association for Computational Linguistics",
    url = "https://aclanthology.org/2023.findings-emnlp.599/",
    doi = "10.18653/v1/2023.findings-emnlp.599",
    pages = "8928--8942"
}

@Article{emirtekin2025llmreview,
AUTHOR = {Emirtekin, Emrah},
TITLE = {Large Language Model-Powered Automated Assessment: A Systematic Review},
JOURNAL = {Applied Sciences},
VOLUME = {15},
YEAR = {2025},
NUMBER = {10},
ARTICLE-NUMBER = {5683},
URL = {https://www.mdpi.com/2076-3417/15/10/5683},
ISSN = {2076-3417},
DOI = {10.3390/app15105683}
}

@inproceedings{lu2024ahp,
    title = "{AHP}-Powered {LLM} Reasoning for Multi-Criteria Evaluation of Open-Ended Responses",
    author = "Lu, Xiaotian  and
      Li, Jiyi  and
      Takeuchi, Koh  and
      Kashima, Hisashi",
    editor = "Al-Onaizan, Yaser  and
      Bansal, Mohit  and
      Chen, Yun-Nung",
    booktitle = "Findings of the Association for Computational Linguistics: EMNLP 2024",
    month = nov,
    year = "2024",
    address = "Miami, Florida, USA",
    publisher = "Association for Computational Linguistics",
    url = "https://aclanthology.org/2024.findings-emnlp.101/",
    doi = "10.18653/v1/2024.findings-emnlp.101",
    pages = "1847--1856"
}

@inproceedings{xie2024dsgram,
author = {Xie, Jinxiang and Li, Yilin and Yin, Xunjian and Wan, Xiaojun},
title = {DSGram: dynamic weighting sub-metrics for grammatical error correction in the era of large language models},
year = {2025},
isbn = {978-1-57735-897-8},
publisher = {AAAI Press},
url = {https://doi.org/10.1609/aaai.v39i24.34746},
doi = {10.1609/aaai.v39i24.34746},
booktitle = {Proceedings of the Thirty-Ninth AAAI Conference on Artificial Intelligence and Thirty-Seventh Conference on Innovative Applications of Artificial Intelligence and Fifteenth Symposium on Educational Advances in Artificial Intelligence},
articleno = {2848},
numpages = {9},
series = {AAAI'25/IAAI'25/EAAI'25}
}

@misc{wu2026doc2ahp,
      title={Doc2AHP: Inferring Structured Multi-Criteria Decision Models via Semantic Trees with LLMs}, 
      author={Hongjia Wu and Shuai Zhou and Hongxin Zhang and Wei Chen},
      year={2026},
      eprint={2601.16479},
      archivePrefix={arXiv},
      primaryClass={cs.AI},
      url={https://arxiv.org/abs/2601.16479}, 
}

@misc{llmAHP2025,
      title={LLM-Assisted AHP for Explainable Cyber Range Evaluation}, 
      author={Vyron Kampourakis and Georgios Kavallieratos and Georgios Spathoulas and Vasileios Gkioulos and Sokratis Katsikas},
      year={2025},
      eprint={2512.10487},
      archivePrefix={arXiv},
      primaryClass={cs.CR},
      url={https://arxiv.org/abs/2512.10487}, 
}

@inproceedings{
tan2025judgebench,
title={JudgeBench: A Benchmark for Evaluating {LLM}-Based Judges},
author={Sijun Tan and Siyuan Zhuang and Kyle Montgomery and William Yuan Tang and Alejandro Cuadron and Chenguang Wang and Raluca Popa and Ion Stoica},
booktitle={The Thirteenth International Conference on Learning Representations},
year={2025},
url={https://openreview.net/forum?id=G0dksFayVq}
}

@misc{wang2024mmluprorobustchallengingmultitask,
      title={MMLU-Pro: A More Robust and Challenging Multi-Task Language Understanding Benchmark}, 
      author={Yubo Wang and Xueguang Ma and Ge Zhang and Yuansheng Ni and Abhranil Chandra and Shiguang Guo and Weiming Ren and Aaran Arulraj and Xuan He and Ziyan Jiang and Tianle Li and Max Ku and Kai Wang and Alex Zhuang and Rongqi Fan and Xiang Yue and Wenhu Chen},
      year={2024},
      eprint={2406.01574},
      archivePrefix={arXiv},
      primaryClass={cs.CL},
      url={https://arxiv.org/abs/2406.01574}, 
}

@misc{liu2023gevalnlgevaluationusing,
      title={G-Eval: NLG Evaluation using GPT-4 with Better Human Alignment}, 
      author={Yang Liu and Dan Iter and Yichong Xu and Shuohang Wang and Ruochen Xu and Chenguang Zhu},
      year={2023},
      eprint={2303.16634},
      archivePrefix={arXiv},
      primaryClass={cs.CL},
      url={https://arxiv.org/abs/2303.16634}, 
}

@misc{white2025livebenchchallengingcontaminationlimitedllm,
      title={LiveBench: A Challenging, Contamination-Limited LLM Benchmark}, 
      author={Colin White and Samuel Dooley and Manley Roberts and Arka Pal and Ben Feuer and Siddhartha Jain and Ravid Shwartz-Ziv and Neel Jain and Khalid Saifullah and Sreemanti Dey and Shubh-Agrawal and Sandeep Singh Sandha and Siddartha Naidu and Chinmay Hegde and Yann LeCun and Tom Goldstein and Willie Neiswanger and Micah Goldblum},
      year={2025},
      eprint={2406.19314},
      archivePrefix={arXiv},
      primaryClass={cs.CL},
      url={https://arxiv.org/abs/2406.19314}, 
}

@misc{openai2025gptoss120bgptoss20bmodel,
      title={gpt-oss-120b \& gpt-oss-20b Model Card}, 
      author={OpenAI},
      year={2025},
      eprint={2508.10925},
      archivePrefix={arXiv},
      primaryClass={cs.CL},
      url={https://arxiv.org/abs/2508.10925}, 
}

@book{dubois1980fuzzy,
  title={Fuzzy Sets and Systems: Theory and Applications},
  author={Dubois, D.J.},
  isbn={9780080917726},
  series={Mathematics in Science and Engineering},
  url={https://books.google.ru/books?id=JmjfHUUtMkMC},
  year={1980},
  publisher={Academic Press}
}

@article{Buckley1985FAHP,
title = {Fuzzy hierarchical analysis},
journal = {Fuzzy Sets and Systems},
volume = {17},
number = {3},
pages = {233-247},
year = {1985},
issn = {0165-0114},
doi = {https://doi.org/10.1016/0165-0114(85)90090-9},
url = {https://www.sciencedirect.com/science/article/pii/0165011485900909},
author = {J.J. Buckley}
}

@book{saaty1980analytic,
  title={The Analytic Hierarchy Process: Planning, Priority Setting, Resource Allocation},
  author={Saaty, T.L.},
  isbn={9780070543713},
  lccn={79041060},
  series={Advanced book program},
  url={https://books.google.ru/books?id=Xxi7AAAAIAAJ},
  year={1980},
  publisher={McGraw-Hill International Book Company}
}

@article{CRAWFORD1985387,
title = {A note on the analysis of subjective judgment matrices},
journal = {Journal of Mathematical Psychology},
volume = {29},
number = {4},
pages = {387-405},
year = {1985},
issn = {0022-2496},
doi = {https://doi.org/10.1016/0022-2496(85)90002-1},
url = {https://www.sciencedirect.com/science/article/pii/0022249685900021},
author = {Gordon Crawford and Cindy Williams}
}

@article{ZESHUI1999443,
title = {A consistency improving method in the analytic hierarchy process1Research supported by NSF of China and Shandong.1},
journal = {European Journal of Operational Research},
volume = {116},
number = {2},
pages = {443-449},
year = {1999},
issn = {0377-2217},
doi = {https://doi.org/10.1016/S0377-2217(98)00109-X},
url = {https://www.sciencedirect.com/science/article/pii/S037722179800109X},
author = {Xu Zeshui and Wei Cuiping}
}

@article{Jos2006ConsistencyIT,
  title={Consistency in the Analytic Hierarchy Process: a New Approach},
  author={Andr{\'e}a Serafim Jos{\'e} and Alonso and Teresa Mª and Lamata},
  journal={Int. J. Uncertain. Fuzziness Knowl. Based Syst.},
  year={2006},
  volume={14},
  pages={445-459},
  url={https://api.semanticscholar.org/CorpusID:18104088}
}

@book{Saaty2012ModelsMC,
  title={Models, Methods, Concepts \& Applications of the Analytic Hierarchy Process},
  author={Saaty, T.L. and Vargas, L.G.},
  isbn={9781461435969},
  series={International Series in Operations Research \& Management Science},
  url={https://books.google.ru/books?id=6J9XI8I1qjwC},
  year={2012},
  publisher={Springer}
}

@misc{li2026gradingscaleimpactllmasajudge,
      title={Grading Scale Impact on LLM-as-a-Judge: Human-LLM Alignment Is Highest on 0-5 Grading Scale}, 
      author={Weiyue Li and Minda Zhao and Weixuan Dong and Jiahui Cai and Yuze Wei and Michael Pocress and Yi Li and Wanyan Yuan and Xiaoyue Wang and Ruoyu Hou and Kaiyuan Lou and Wenqi Zeng and Yutong Yang and Yilun Du and Mengyu Wang},
      year={2026},
      eprint={2601.03444},
      archivePrefix={arXiv},
      primaryClass={cs.CL},
      url={https://arxiv.org/abs/2601.03444}, 
}

@misc{qwen3.5,
    title  = {{Qwen3.5}: Towards Native Multimodal Agents},
    author = {{Qwen Team}},
    month  = {February},
    year   = {2026},
    url    = {https://qwen.ai/blog?id=qwen3.5}
}

\clearpage

\appendix

\section{Full Results}
\label{sec:appendix_results}

\begin{sideways}
\begin{minipage}[c][][c]{0.9\textheight}

\centering
\setlength{\tabcolsep}{1pt}

\begin{threeparttable}

\caption{
Performance comparison between \textbf{gpt-oss-20b} and
\textbf{gpt-oss-120b} across dataset splits under two scoring
scales. Values denote accuracy (\%). Higher is better.
}
\fontsize{7}{7}\selectfont 
\label{tab:gpt_combined_scales}

\begin{tabular}{
llc
*{5}{S}*{5}{S}||
*{5}{S}*{5}{S}
}
\toprule

\multirow{4}{*}{\textbf{Category}} &
\multirow{4}{*}{\textbf{Split}} &
\multirow{4}{*}{\textbf{N}} &
\multicolumn{10}{c}{\textbf{gpt-oss-20b}} &
\multicolumn{10}{c}{\textbf{gpt-oss-120b}} \\

\cmidrule(lr){4-13}
\cmidrule(lr){14-23}

&&&
\multicolumn{5}{c}{1--10 Scale} &
\multicolumn{5}{c}{1--5 Scale} &
\multicolumn{5}{c}{1--10 Scale} &
\multicolumn{5}{c}{1--5 Scale} \\

\cmidrule(lr){4-8}
\cmidrule(lr){9-13}
\cmidrule(lr){14-18}
\cmidrule(lr){19-23}

&&&
{Dir} & {Cri} & {Fuz} & {DC} & {DF} &
{Dir} & {Cri} & {Fuz} & {DC} & {DF} &
{Dir} & {Cri} & {Fuz} & {DC} & {DF} &
{Dir} & {Cri} & {Fuz} & {DC} & {DF} \\

\midrule

\multicolumn{23}{l}{\textbf{LiveBench}}\\
\addlinespace[2pt]

\multirow{2}{*}{math}
& GPT & 56
& 82.1 & 87.5 &  89.3 & 87.5 & 87.5
& 75.0 & 87.5 &  89.3 & 87.5 & 87.5
& 87.5 & 89.3 & 89.3 &  92.9 &  92.9
& 82.1 & 85.7 & 85.7 &  87.5 &  87.5 \\

& Claude & 34
& 67.6 & 79.4 &  91.2 & 76.5 &  91.2
& 79.4 & 79.4 & 79.4 & 79.4 & 79.4
& 76.5 &  82.4 &  82.4 &  82.4 &  82.4
& 76.5 &  82.4 &  82.4 &  82.4 &  82.4 \\

\addlinespace

\multirow{2}{*}{reasoning}
& GPT & 98
& 73.5 & 73.5 & 73.5 &  77.6 &  77.6
& 80.6 & 85.7 & 85.7 &  88.8 &  88.8
& 88.8 & 90.8 &  91.8 &  91.8 &  91.8
& 87.8 & 87.8 & 87.8 &  88.8 &  88.8 \\

& Claude & 51
& 80.4 & 80.4 & 78.4 &  88.2 & 86.3
&  86.3 & 70.6 & 76.5 & 82.4 & 82.4
& 84.3 &  94.1 &  94.1 &  94.1 &  94.1
& 88.2 & 90.2 & 90.2 &  92.2 & 90.2 \\

\addlinespace

\multirow{2}{*}{livecodebench}
& GPT & 42
& 83.3 & 88.1 & 90.5 & 92.9 &  95.2
& 85.7 & 83.3 & 83.3 &  92.9 &  92.9
& 92.9 & 90.5 & 90.5 &  95.2 &  95.2
& 90.5 & 95.2 & 95.2 &  97.6 &  97.6 \\

& Claude & 31
& 74.2 & 74.2 & 71.0 &  90.3 &  90.3
& 71.0 & 80.6 & 80.6 &  87.1 &  87.1
& 67.7 & 83.9 & 87.1 &  90.3 &  90.3
& 74.2 &  80.6 &  80.6 &  80.6 &  80.6 \\

\midrule
\multicolumn{23}{l}{\textbf{MMLU-Pro}}\\

\multirow{2}{*}{biology} & GPT & 11 & 54.5 & 63.6 & 63.6 & 63.6 & 63.6 & 54.5 & 63.6 & 63.6 & 63.6 & 63.6 & 54.5 & 45.5 & 45.5 & 45.5 & 45.5 & 54.5 & 54.5 & 54.5 & 54.5 & 54.5 \\
& Claude & 11 & 54.5 & 63.6 & 63.6 & 54.5 & 63.6 & 45.5 & 63.6 & 63.6 & 63.6 & 63.6 & 54.5 & 63.6 & 63.6 & 63.6 & 63.6 & 54.5 & 63.6 & 63.6 & 63.6 & 63.6 \\

\multirow{2}{*}{business} & GPT & 11 & 72.7 & 63.6 & 63.6 & 63.6 & 63.6 & 81.8 & 54.5 & 54.5 & 63.6 & 63.6 & 90.9 & 63.6 & 63.6 & 81.8 & 81.8 & 72.7 & 81.8 & 81.8 & 81.8 & 81.8 \\
& Claude & 11 & 72.7 & 72.7 & 72.7 & 72.7 & 72.7 & 72.7 & 72.7 & 72.7 & 72.7 & 72.7 & 54.5 & 72.7 & 72.7 & 72.7 & 72.7 & 54.5 & 63.6 & 72.7 & 63.6 & 63.6 \\

\multirow{2}{*}{chemistry} & GPT & 11 & 63.6 & 81.8 & 81.8 & 81.8 & 81.8 & 90.9 & 81.8 & 90.9 & 90.9 & 90.9 & 81.8 & 90.9 & 90.9 & 90.9 & 90.9 & 72.7 & 81.8 & 81.8 & 81.8 & 81.8 \\
& Claude & 11 & 81.8 & 81.8 & 81.8 & 90.9 & 81.8 & 81.8 & 81.8 & 81.8 & 81.8 & 81.8 & 81.8 & 90.9 & 90.9 & 90.9 & 90.9 & 90.9 & 81.8 & 81.8 & 81.8 & 81.8  \\

\multirow{2}{*}{CS} & GPT & 11 & 63.6 & 81.8 & 81.8 & 81.8 & 81.8 & 72.7 & 72.7 & 72.7 & 72.7 & 72.7 & 72.7 & 81.8 & 81.8 & 81.8 & 81.8 & 63.6 & 81.8 & 81.8 & 81.8 & 81.8 \\
& Claude & 11 & 72.7 & 72.7 & 72.7 & 72.7 & 72.7 & 72.7 & 72.7 & 72.7 & 81.8 & 81.8 & 81.8 & 81.8 & 81.8 & 81.8 & 81.8 & 90.9 & 81.8 & 81.8 & 90.9 & 90.9 \\

\multirow{2}{*}{economics} & GPT & 11 & 54.5 & 63.6 & 63.6 & 72.7 & 72.7 & 63.6 & 54.5 & 54.5 & 54.5 & 54.5 & 63.6 & 63.6 & 63.6 & 72.7 & 72.7 & 63.6 & 54.5 & 54.5 & 54.5 & 54.5 \\
& Claude & 11 & 54.5 & 54.5 & 54.5 & 63.6 & 63.6 & 54.5 & 54.5 & 54.5 & 54.5 & 54.5 & 63.6 & 72.7 & 63.6 & 63.6 & 63.6 & 72.7 & 81.8 & 81.8 & 81.8 & 72.7 \\

\multirow{2}{*}{engineering} & GPT & 11 & 72.7 & 63.6 & 63.6 & 63.6 & 63.6 & 45.5 & 81.8 & 81.8 & 81.8 & 81.8 & 72.7 & 90.9 & 90.9 & 100.0 & 100.0 & 72.7 & 81.8 & 81.8 & 81.8 & 81.8 \\
& Claude & 11 & 45.5 & 63.6 & 63.6 & 63.6 & 63.6 & 36.4 & 54.5 & 54.5 & 54.5 & 54.5 & 54.5 & 81.8 & 81.8 & 81.8 & 81.8 & 63.6 & 81.8 & 81.8 & 81.8 & 81.8 \\

\multirow{2}{*}{health} & GPT & 11 & 54.5 & 72.7 & 72.7 & 72.7 & 72.7 & 63.6 & 90.9 & 90.9 & 90.9 & 90.9 & 54.5 & 81.8 & 81.8 & 72.7 & 72.7 & 54.5 & 63.6 & 63.6 & 63.6 & 63.6 \\
& Claude & 11 & 90.9 & 72.7 & 72.7 & 81.8 & 81.8 & 81.8 & 72.7 & 72.7 & 72.7 & 72.7 & 72.7 & 81.8 & 81.8 & 81.8 & 81.8 & 81.8 & 81.8 & 81.8 & 81.8 & 81.8 \\

\multirow{2}{*}{history} & GPT & 11 & 81.8 & 90.9 & 90.9 & 90.9 & 90.9 & 81.8 & 81.8 & 81.8 & 81.8 & 81.8 & 72.7 & 54.5 & 54.5 & 63.6 & 63.6 & 72.7 & 63.6 & 63.6 & 72.7 & 72.7 \\
& Claude & 11 & 54.5 & 36.4 & 36.4 & 45.5 & 45.5 & 63.6 & 45.5 & 45.5 & 45.5 & 45.5 & 63.6 & 63.6 & 63.6 & 72.7 & 72.7 & 63.6 & 63.6 & 63.6 & 54.5 & 54.5 \\

\multirow{2}{*}{law} & GPT & 11 & 54.5 & 54.5 & 54.5 & 63.6 & 63.6 & 63.6 & 63.6 & 63.6 & 63.6 & 72.7 & 54.5 & 54.5 & 63.6 & 45.5 & 45.5 & 54.5 & 45.5 & 45.5 & 45.5 & 54.5 \\
& Claude & 11 & 36.4 & 72.7 & 72.7 & 72.7 & 72.7 & 36.4 & 54.5 & 54.5 & 45.5 & 54.5 & 45.5 & 72.7 & 72.7 & 72.7 & 72.7 & 54.5 & 63.6 & 63.6 & 63.6 & 63.6 \\

\multirow{2}{*}{math} & GPT & 11 & 90.9 & 90.9 & 90.9 & 90.9 & 90.9 & 81.8 & 90.9 & 90.9 & 90.9 & 90.9 & 81.8 & 81.8 & 81.8 & 81.8 & 81.8 & 90.9 & 81.8 & 81.8 & 81.8 & 81.8 \\
& Claude & 11 & 81.8 & 100.0 & 100.0 & 90.9 & 90.9 & 90.9 & 90.9 & 90.9 & 81.8 & 81.8 & 100.0 & 100.0 & 100.0 & 100.0 & 100.0 & 100.0 & 90.9 & 90.9 & 90.9 & 90.9  \\

\multirow{2}{*}{other} & GPT & 11 & 72.7 & 63.6 & 63.6 & 72.7 & 63.6 & 63.6 & 63.6 & 63.6 & 72.7 & 72.7 & 45.5 & 54.5 & 54.5 & 54.5 & 54.5 & 54.5 & 63.6 & 63.6 & 63.6 & 63.6 \\
& Claude & 11 & 54.5 & 72.7 & 72.7 & 63.6 & 63.6 & 72.7 & 81.8 & 81.8 & 81.8 & 81.8 & 54.5 & 63.6 & 63.6 & 63.6 & 63.6 & 45.5 & 45.5 & 45.5 & 45.5 & 45.5 \\

\multirow{2}{*}{philosophy} & GPT & 11 & 72.7 & 54.5 & 54.5 & 63.6 & 63.6 & 45.5 & 63.6 & 63.6 & 63.6 & 63.6 & 63.6 & 72.7 & 72.7 & 72.7 & 72.7 & 72.7 & 63.6 & 63.6 & 63.6 & 63.6 \\
& Claude & 11 & 27.3 & 54.5 & 54.5 & 36.4 & 36.4 & 45.5 & 45.5 & 45.5 & 45.5 & 45.5 & 72.7 & 45.5 & 45.5 & 45.5 & 45.5 & 45.5 & 45.5 & 45.5 & 45.5 & 45.5  \\

\multirow{2}{*}{physics} & GPT & 11 & 36.4 & 63.6 & 63.6 & 63.6 & 63.6 & 45.5 & 90.9 & 90.9 & 81.8 & 81.8 & 72.7 & 81.8 & 81.8 & 81.8 & 81.8 & 36.4 & 72.7 & 72.7 & 72.7 & 72.7 \\
& Claude & 11 & 100.0 & 90.9 & 90.9 & 90.9 & 90.9 & 81.8 & 90.9 & 100.0 & 90.9 & 100.0 & 100.0 & 90.9 & 100.0 & 100.0 & 100.0 & 81.8 & 90.9 & 90.9 & 90.9 & 90.9 \\

\multirow{2}{*}{psychology} & GPT & 11 & 45.5 & 72.7 & 81.8 & 63.6 & 63.6 & 45.5 & 63.6 & 63.6 & 63.6 & 63.6 & 54.5 & 63.6 & 63.6 & 54.5 & 54.5 & 36.4 & 63.6 & 63.6 & 63.6 & 63.6 \\
& Claude & 11 & 36.4 & 45.5 & 45.5 & 36.4 & 36.4 & 36.4 & 27.3 & 27.3 & 27.3 & 27.3 & 27.3 & 36.4 & 36.4 & 36.4 & 36.4 & 36.4 & 54.5 & 27.3 & 27.3 & 27.3  \\

\midrule

\multirow{2}{*}{\textbf{Overall}}
& GPT & 350
& 71.7 & 76.0 & 76.9 &  78.6 &  78.6
& 73.1 & 80.0 & 80.6 & 82.6 &  82.9
& 79.4 & 81.4 & 82.0 &  83.4 &  83.4
& 76.0 & 79.7 & 79.7 & 80.9 &  81.1 \\

& Claude & 270
& 67.4 & 72.6 & 73.3 & 74.8 &  76.3
& 70.0 & 69.6 & 71.1 & 72.2 &  73.0
& 71.1 & 79.3 & 79.6 &  80.4 &  80.4
& 73.7 & 75.9 &  76.3 &  76.3 & 75.6 \\

\bottomrule
\end{tabular}

\end{threeparttable}
\end{minipage}
\end{sideways}


\begin{sideways}
\begin{minipage}[c][][c]{1\textheight}

\centering
\setlength{\tabcolsep}{1pt}

\begin{threeparttable}

\caption{
Performance comparison between \textbf{Qwen3.5-9B} and
\textbf{Qwen3.5-35B-A3B} across dataset splits under two scoring
scales. Values denote accuracy (\%). Higher is better.
}
\fontsize{7}{7}\selectfont 
\label{tab:gpt_combined_scales}

\begin{tabular}{
llc
*{5}{S}*{5}{S}||
*{5}{S}*{5}{S}
}
\toprule

\multirow{4}{*}{\textbf{Category}} &
\multirow{4}{*}{\textbf{Split}} &
\multirow{4}{*}{\textbf{N}} &
\multicolumn{10}{c}{\textbf{Qwen3.5-9B}} &
\multicolumn{10}{c}{\textbf{Qwen3.5-35B-A3B}} \\

\cmidrule(lr){4-13}
\cmidrule(lr){14-23}

&&&
\multicolumn{5}{c}{1--10 Scale} &
\multicolumn{5}{c}{1--5 Scale} &
\multicolumn{5}{c}{1--10 Scale} &
\multicolumn{5}{c}{1--5 Scale} \\

\cmidrule(lr){4-8}
\cmidrule(lr){9-13}
\cmidrule(lr){14-18}
\cmidrule(lr){19-23}

&&&
{Dir} & {Cri} & {Fuz} & {DC} & {DF} &
{Dir} & {Cri} & {Fuz} & {DC} & {DF} &
{Dir} & {Cri} & {Fuz} & {DC} & {DF} &
{Dir} & {Cri} & {Fuz} & {DC} & {DF} \\

\midrule

\multicolumn{23}{l}{\textbf{LiveBench}}\\
\addlinespace[2pt]

\multirow{2}{*}{math}
& GPT & 56&
91.1& 87.5& 89.3& 87.5& 87.5&
87.5& 82.1& 82.1& 89.3& 87.5&
89.3& 82.1& 83.9& 87.5& 87.5&
83.9& 87.5& 89.3& 89.3& 89.3

\\

& Claude & 34
&85.3& 85.3& 88.2& 88.2& 88.2&
91.2& 88.2& 91.2& 94.1& 94.1&
91.2& 88.2& 91.2& 88.2& 88.2&
94.1& 79.4& 85.3& 94.1& 94.1

\\

\addlinespace

\multirow{2}{*}{reasoning}
& GPT & 98
& 95.9& 90.8& 91.8& 92.9& 93.9&
94.9& 89.8& 89.8& 90.8& 89.8&
93.9& 85.7& 89.8& 91.8& 91.8&
93.9& 89.8& 93.9& 92.9& 92.9

\\

& Claude & 51
&88.2& 88.2& 92.2& 92.2& 94.1&
88.2& 88.2& 90.2& 92.2& 92.2&
96.1& 96.1& 96.1& 98.0& 98.0&
100.0& 92.2& 92.2& 92.2& 92.2

 \\

\addlinespace

\multirow{2}{*}{livecodebench}
& GPT & 42
& 78.6& 95.2& 97.6& 95.2& 95.2&
73.8& 88.1& 85.7& 85.7& 83.3&
92.9& 85.7& 88.1& 92.9& 92.9&
85.7& 92.9& 92.9& 92.9& 92.9

\\

& Claude & 31
&58.1& 77.4& 77.4& 80.6& 80.6&
67.7& 80.6& 83.9& 83.9& 83.9&
77.4& 87.1& 87.1& 93.5& 90.3&
87.1& 90.3& 90.3& 90.3& 90.3

\\

\midrule
\multicolumn{23}{l}{\textbf{MMLU-Pro}}\\

\multirow{2}{*}{biology} & GPT & 11 & 63.6& 54.5& 54.5& 54.5& 54.5& 
63.6& 72.7& 72.7& 72.7& 72.7&72.7& 81.8& 81.8& 72.7& 72.7&
72.7& 63.6& 72.7& 72.7& 72.7

\\

& Claude & 11 & 
63.6& 45.5& 45.5& 54.5& 54.5&
54.5& 63.6& 63.6& 63.6& 63.6&
54.5& 63.6& 63.6& 63.6& 63.6&
54.5& 72.7& 72.7& 72.7& 72.7

\\

\multirow{2}{*}{business} & GPT & 11 & 81.8& 63.6& 72.7& 72.7& 63.6&
72.7& 54.5& 63.6& 63.6& 63.6&
81.8& 63.6& 63.6& 63.6& 63.6&
81.8& 72.7& 72.7& 72.7& 72.7

\\
& Claude & 11 &  90.9& 72.7& 81.8& 81.8& 81.8&
90.9& 90.9& 90.9& 90.9& 90.9&
90.9& 90.9& 90.9& 90.9& 90.9&
100.0& 90.9& 90.9& 90.9& 90.9

\\

\multirow{2}{*}{chemistry} & GPT & 11 & 81.8& 90.9& 100.0& 100.0& 100.0&
90.9& 90.9& 90.9& 90.9& 90.9&
90.9& 81.8& 81.8& 81.8& 81.8&
81.8& 72.7& 72.7& 72.7& 72.7

\\
& Claude & 11 & 90.9& 72.7& 72.7& 81.8& 81.8&
90.9& 90.9& 90.9& 90.9& 90.9&
90.9& 81.8& 81.8& 81.8& 81.8&
90.9& 77.3& 77.3& 81.8& 81.8

\\

\multirow{2}{*}{CS} & GPT & 11 & 63.6& 72.7& 72.7& 72.7& 72.7&
63.6& 72.7& 72.7& 72.7& 72.7&
90.9& 90.9& 90.9& 90.9& 90.9&
90.9& 90.9& 90.9& 81.8& 81.8

\\
& Claude & 11 &100.0& 100.0& 100.0& 100.0& 100.0&
81.8& 90.9& 90.9& 90.9& 90.9&
 90.9& 90.9& 90.9& 90.9& 90.9&
81.8& 81.8& 90.9& 90.9& 90.9

\\

\multirow{2}{*}{economics} & GPT & 11 & 90.9& 90.9& 90.9& 90.9& 90.9&
81.8& 100.0& 100.0& 100.0& 100.0&
100.0& 90.9& 90.9& 100.0& 100.0&
100.0& 90.9& 100.0& 100.0& 100.0

\\
& Claude & 11 & 72.7& 72.7& 81.8& 72.7& 72.7&
72.7& 81.8& 81.8& 81.8& 81.8&
81.8& 81.8& 81.8& 81.8& 81.8&
81.8& 81.8& 81.8& 81.8& 81.8

\\

\multirow{2}{*}{engineering} & GPT & 11 & 100.0& 100.0& 100.0& 100.0& 100.0& 
100.0& 81.8& 81.8& 81.8& 81.8&
100.0& 81.8& 100.0& 100.0& 100.0&
100.0& 100.0& 100.0& 100.0& 100.0

\\
& Claude & 11 & 90.9& 90.9& 100.0& 90.9& 90.9&
81.8& 81.8& 90.9& 90.9& 90.9&
90.9& 72.7& 72.7& 72.7& 72.7&
72.7& 90.9& 90.9& 90.9& 81.8

\\

\multirow{2}{*}{health} & GPT & 11 & 63.6& 72.7& 72.7& 72.7& 72.7&
54.5& 63.6& 63.6& 54.5& 54.5&
 54.5& 54.5& 54.5& 54.5& 54.5&
54.5& 63.6& 63.6& 63.6& 63.6

\\
& Claude & 11 &90.9& 72.7& 72.7& 72.7& 72.7&
90.9& 90.9& 100.0& 90.9& 90.9&
81.8& 72.7& 81.8& 81.8& 81.8&
86.4& 81.8& 81.8& 86.4& 86.4

\\

\multirow{2}{*}{history} & GPT & 11 & 90.9& 72.7& 72.7& 81.8& 81.8&
81.8& 90.9& 90.9& 90.9& 90.9&
 81.8& 90.9& 90.9& 90.9& 90.9&
100.0& 100.0& 100.0& 100.0& 100.0

\\
& Claude & 11 &              72.7& 45.5& 54.5& 45.5& 45.5&
54.5& 72.7& 72.7& 72.7& 72.7&
81.8& 63.6& 63.6& 72.7& 72.7&
52.6& 52.6& 63.2& 63.2& 63.2

\\

\multirow{2}{*}{law} & GPT & 11 & 81.8& 90.9& 90.9& 90.9& 90.9&
81.8& 63.6& 72.7& 72.7& 72.7&
90.9& 72.7& 72.7& 81.8& 81.8&
90.9& 81.8& 90.9& 90.9& 90.9

\\
& Claude & 11 &                  100.0& 81.8& 81.8& 81.8& 81.8&
 81.8& 100.0& 100.0& 90.9& 90.9&
90.9& 90.9& 100.0& 90.9& 100.0&
81.8& 81.8& 90.9& 90.9& 90.9

\\

\multirow{2}{*}{math} & GPT & 11 & 90.9& 90.9& 90.9& 90.9& 90.9&
90.9& 81.8& 90.9& 90.9& 90.9&
90.9& 90.9& 90.9& 90.9& 90.9&
90.9& 81.8& 90.9& 90.9& 90.9

\\
& Claude & 11 &                100.0& 100.0& 100.0& 100.0& 100.0&
 100.0& 90.9& 100.0& 100.0& 100.0&
 100.0& 100.0& 100.0& 100.0& 100.0&
100.0& 95.5& 100.0& 100.0& 100.0

   \\

\multirow{2}{*}{other} & GPT & 11 & 54.5& 63.6& 63.6& 63.6& 63.6&
81.8& 81.8& 81.8& 81.8& 81.8&
72.7& 81.8& 81.8& 81.8& 81.8&
72.7& 90.9& 90.9& 90.9& 90.9

\\
& Claude & 11 &  54.5& 54.5& 63.6& 63.6& 63.6&
72.7& 72.7& 72.7& 72.7& 72.7&
 63.6& 81.8& 81.8& 81.8& 81.8&
63.6& 72.7& 81.8& 81.8& 81.8

\\

\multirow{2}{*}{philosophy} & GPT & 11 & 81.8& 90.9& 90.9& 100.0& 90.9& 
81.8& 72.7& 72.7& 72.7& 72.7&90.9& 72.7& 72.7& 81.8& 81.8&
81.8& 81.8& 81.8& 81.8& 81.8

\\
& Claude & 11 &54.5& 45.5& 45.5& 54.5& 54.5&
54.5& 36.4& 36.4& 36.4& 36.4&
 81.8& 90.9& 90.9& 90.9& 90.9&
 90.5& 57.1& 61.9& 76.2& 76.2

\\

\multirow{2}{*}{physics} & GPT & 11 & 90.9& 90.9& 90.9& 90.9& 90.9&
90.9& 100.0& 100.0& 100.0& 100.0&100.0& 90.9& 90.9& 90.9& 90.9&
 90.9& 100.0& 100.0& 100.0& 100.0

\\
& Claude & 11 &     100.0& 90.9& 90.9& 100.0& 100.0&
100.0& 90.9& 90.9& 90.9& 90.9&
100.0& 100.0& 100.0& 100.0& 100.0&
 90.9& 90.9& 90.9& 100.0& 90.9

  \\

\multirow{2}{*}{psychology} & GPT & 11 & 54.5& 45.5& 45.5& 45.5& 45.5&
54.5& 63.6& 63.6& 72.7& 72.7&
54.5& 63.6& 63.6& 72.7& 72.7&
 54.5& 54.5& 54.5& 54.5& 54.5
\\
& Claude & 11 & 36.4& 45.5& 45.5& 45.5& 45.5&
45.5& 36.4& 45.5& 45.5& 45.5&
63.6& 54.5& 54.5& 54.5& 54.5&
54.5& 59.1& 59.1& 54.5& 54.5

\\

\midrule

\multirow{2}{*}{\textbf{Overall}}
& GPT & 350
&85.1&85.1&86.6&86.9&86.6&
83.7&83.1&83.7&85.1&84.3&
88.6&82.3&84.6&87.1&87.1&
86.6&86.3&88.9&88.3&88.3
\\

& Claude & 270&
79.6&76.7&79.6&80.4&80.7&
79.6&81.5&84.1&84.1&84.1&
85.8&85.4&86.6&87.3&87.3&
84.6&80.1&82.8&85.2&84.6

 \\

\bottomrule
\end{tabular}

\end{threeparttable}
\end{minipage}
\end{sideways}

\end{document}